\documentclass{article}

\usepackage{times}
\usepackage{booktabs}
\usepackage{hyperref}
\usepackage{graphicx}
\usepackage{url}
\usepackage{subfiles}
\usepackage{multirow}
\usepackage{longtable}
\usepackage{natbib}
\usepackage{csquotes}
\usepackage{amsmath,amsfonts,bm,amsthm}
\usepackage{}
\usepackage{mathtools,algorithm,algorithmic}
\usepackage{tikz}
\usetikzlibrary{arrows.meta, positioning}

\newcommand{\xhdr}[1]{\vspace{2mm}\noindent{\textbf{ #1.}}}

\newcommand{\E}{\mathbb{E}}

\newcommand{\R}{\mathbb{R}}

\newcommand{\str}{\text{str}}

\newtheorem{theorem}{Theorem}[section]

\newtheorem{lemma}{Lemma}[section]

\newcommand{\ommmit}[1]{}
\newcommand{\background}[1]{}

\title{Equitable Evaluation via Elicitation\thanks{This work was supported in part by the Hire Aspirations Institute, grant G-2017-9890 from the Alfred P.\ Sloan Foundation, and the Simons Collaboration on Theory of Algorithmic Fairness. Authors are listed alphabetically.}}

\author{
    Elbert Du \\
    \textit{Harvard University}
    \and
    Cynthia Dwork \\
    \textit{Harvard University}
    \and
    Lunjia Hu \\
    \textit{Northeastern University}
    \and
    Reid McIlroy-Young \\
    \textit{Harvard University}
    \and
    Han Shao \\
    \textit{University of Maryland}
    \and
    Linjun Zhang \\
    \textit{Rutgers University}
}
\date{January 2026}

\begin{document}

\maketitle

\begin{abstract}
Individuals with similar qualifications and skills may vary in their demeanor, or outward manner: some tend toward self-promotion while others are modest to the point of omitting crucial information.  Comparing the self-descriptions of equally qualified job-seekers with different self-presentation styles is therefore problematic.

We build an interactive AI for \textit{skill elicitation} that provides accurate determination of skills \textit{while simultaneously allowing individuals to speak in their own voice.} Such a system can be deployed, for example, when a new user joins a professional networking platform, or when matching employees to needs during a company reorganization.  To obtain sufficient training data, we train an LLM to act as synthetic humans.

Elicitation mitigates {\em endogenous} bias arising from individuals’ own self‑reports.  To address systematic {\em model} bias we enforce a mathematically rigorous notion of equitability ensuring that the covariance between self-presentation manner and skill evaluation error is small.

\end{abstract}

\section{Introduction}

It is widely known that biases in data yield biased systems.  Self-descriptions are not exempt from bias: data about individuals that are provided by the individuals themselves exhibit cultural biases.
More broadly, individuals, even within the same culture, differ in their levels of modesty and tendency (or ability) to self-promote.  These differences have measurable downstream effects on outcomes, such as recruitment~\cite{murciano2022missing}.

If the problem were simply one of tone, existing language models could mitigate the effect by homogenizing the presentation of facts supplied by the individuals, for example, by using a prompt of the form ``Rewrite this in the style of a white cisgender male in an individualistic culture.''  However, the LLM cannot compensate for relevant information that the user {\em fails to supply}, which studies show to exhibit pronounced cultural and gender disparities~\citep{LinkedIn2017,murciano2022missing,deschacht2017cross}
\footnote{For example, in their official blog, LinkedIn reports~\cite{LinkedIn2017}: ``In the U.S., women on average include 11\% less skills than men on their LinkedIn profile, {\em even at similar occupations and experience levels}'' (emphasis added; see also~\cite{ahuja2024right}). This number is matched in Murciano-Goroff's brilliant measurement of the gender gap in self-reporting of skills via individuals' actual previous coding work as seen on GitHub and taking into account ``the number of lines of code that a candidate has contributed to \textit{others'} open source projects as an indication of the reputation of the candidate's coding work'' (emphasis added)~\cite{murciano2022missing}. There is also cross-cultural research finding a self-promotion gap between collectivist and individualist cultures; see~\cite{deschacht2017cross}, who find such a gap (and also find the gender gap firmly persistent across different cultures).}

These {\em endogenous biases} arising from individuals' self presentation styles cannot be addressed by post-hoc fairness techniques such as multicalibration~\cite{hebert2018multicalibration}, outcome indistinguishability~\cite{dwork2021outcome}, and learning fair representations~\cite{zemel2013learning} because the necessary data may simply not be present, or because the semantics of identical text for people with different styles may differ.

For concreteness (and availability of data), we focus on job-seekers' self-generated professional profiles, for example, as on LinkedIn, but our proof of concept can extend to many other settings.  Our key contribution is an interactive tool for accurate and equitable elicitation of skill signals in job candidates across a cultural spectrum.  That is, we determine whether or not a person has a specific skill based on their initial self-description and answers to a few questions.
Crucially -- and this is a principal design goal -- our tool allows individuals to speak in the style in which they are comfortable.  The pressure, and even well-intentioned advice, to conform to a specific interaction style is widespread, burdening even highly competent but conversationally more restrained individuals~\cite{mohr2014women,ahuja2024right}.  To our knowledge, ours is the first work to actively elicit information while respecting personal style.

 For our purposes, a \textit{predictor} is a function that maps a transcript to a score in $[0,1]$.  Speaking intuitively and informally, one can think of this score as a ``probability'' that the individual in the transcript has a stated skill.  For our purposes, it is helpful to think of the algorithm as trying to learn a ground truth, rather than to predict the future~\footnote{The meaning of a probability for a non-repeatable event is an unresolved theoretical question. For more on this in the context of the predictor technology used herein, see~\cite{dwork2021outcome}.}.  We use the terms {\em model} and {\em predictor} interchangeably.

Elicitation mitigates self-presentation bias.  To avoid systematic {\em model} biases we impose multi-accuracy~\cite{hebert2018multicalibration} with respect to the Big~5 personality types, which roughly means that the covariance between the personality type and the prediction error is small~\cite{gopalan2021omnipredictors}.  The choice of Big~5 was expedient, as algorithms for scoring these are readily available in the NLP literature.   Indeed, multi-accuracy is completely general and multi-accurate learning algorithms provide the low-covariance guarantee with respect to any pre-specified collection of functions mapping individuals to a finite range~\cite{hebert2018multicalibration,dwork2021outcome,gopalan2022loss}.  In the same vein, we chose multi-accuracy, rather than, for example, the stronger condition of {\em multicalibration}~\cite{hebert2018multicalibration}, for simplicity in this first proof of concept.

Multi-accuracy, which curtails the correlation between prediction error and self-presentation style, can be enforced at any point, including on the individuals' initial self-descriptions.  In our system, the trainings of the {\em information elicitor} and the {\em skill predictor} are interleaved, and so the distribution on conversations in early rounds of training differs from that in later rounds of training. We train in a series of  time periods, and we impose multi-accuracy in each one, see figure~\ref{fig:acc}. In this way, our deployment of multi-accuracy diverges from that in previous work.

Beyond articulating the problem of equitable elicitation while preserving individual voice, our work has several notable features:
\begin{enumerate}
    \item We provide a novel example of the use of LLMs for computational social science research. Drawing on a publicly available dataset of individual profiles, we build an LLM to create synthetic humans on which we train an accurate and equitable scoring function for skills.\footnote{This is the only use of LLMs in our work; had we sufficient data, LLMs would not have been necessary. See the discussion in Section~\ref{sec:discussion}.}
    \item In many commercial settings, the training of predictors is a dynamic process, with systems trained to improve themselves in a sequence of epochs. Let $p_i$ denote the predictor in the $i$th epoch. Our system ensures that every one of these epochal predictors $p_i$ satisfies our model fairness constraints.
\end{enumerate}
To our knowledge, our system is the first in the algorithmic fairness literature to address inadequacies in the data by actively eliciting more information.  We see this as a crucial subject for further study.

\subsection{High-Level Overview}

 We use a structured approach to building an elicitation model based on \cite{wangadaptive} and Sutton \textit{et al.} style reinforcement learning~\cite{sutton1999policy}.

We determine whether or not a person has a specific skill based on their initial self-descriptions and their answers to a series of adaptively selected questions. The question selection algorithm is trained to minimize uncertainty of the predictions, with the questions belonging to a prespecified and human-created set: no user will ever be asked a question that was directly generated by an LLM, and the pool of possible questions can be audited beforehand by any relevant parties~\footnote{These aspects are relevant to deployment in the EU.}

Given a question-and-answer conversation prefix, the selection of the next question is done via a fixed machine-learning-based function, which maps the transcript of user responses to a probability distribution on the next question. This \textit{question selector} converts the user's responses into vectors using an embedding model and then feeds them into our pre-trained transformer-based machine learning system. We use a transformer architecture because that allows us to input a variable-length conversation prefix; this is needed because, for example, on the first question we only have the user's background information or profile, while on the fifth we also have 4 previous answers. This system is static and reproducible: given the same answers, it will produce the same distribution

on outputs (questions).

We construct our \textit{skill predictor} using a technique similar to the one used to build the question selector, and the two functions are trained together (see Section~\ref{sec:methods}).  We formulate the skill prediction task as a scoring function that deterministically maps a transcript of questions and responses to a score in $[0,1]$.
Training proceeds in a series of rounds and multi-accuracy is imposed in each one using a technique

from~\cite{gopalan2022loss}.\footnote{As noted above, stronger guarantees, such as multi-calibration and moment multicalibration~\cite{jung2021moment} can also be ensured, but these algorithms are more complicated.}

We remark that neither of the two functions is generative.
The training process for our elicitation model
uses generative AI (specifically, our trained LLM) to bootstrap and expand the training data, but no generative processes are deployed at runtime.

\section{Related Work}
\subsection{Algorithmic Fairness}
The literature on algorithmic fairness is broad and rapidly growing, encompassing notions such as individual (aka metric) fairness \citep{dwork2012fairness} and group fairness \citep{dwork2012fairness, hardt2016equality}.
Protections defined with respect to a single group can be too weak \citep{dwork2012fairness}, and desirable group-based fairness desiderata can be mutually inconsistent for imperfect classifiers~\cite{chouldechova2017fair} and scoring functions~\citep{kleinberg2018algorithmic}, while finding a metric for individual fairness can be problematic (but see~\citep{ilvento2019metric,mukherjee2020two,rothblum_probably_2018} for positive results).
Bracketing the consistency concerns, the literature explored fixing a single group-fairness condition and requiring that it hold simultaneously for a rich collection of arbitrarily overlapping (sufficiently large) groups
\citep{hebert2018multicalibration,kearns2018preventing,kim2019multiaccuracy}.
Most pertinent to our work are the frameworks of multi-calibration and multi-accuracy \citep{hebert2018multicalibration,kim2019multiaccuracy}, which require predictors to be, respectively, accurate {\em in expectation} or calibrated simultaneously on each group in the collection.
By enforcing multi-accuracy with respect to presentation styles, we extend these fairness-as-accuracy goals into the domain of \emph{information elicitation}, adding proactive interaction to ensure that stylistic self-presentation variation does not systematically affect errors in skill estimation.

A wide range of related approaches highlight complementary fairness notions.
Fair representation learning methods aim to construct latent spaces invariant to sensitive attributes \cite{zemel2013learning,wang_learning_nodate}.

The problem of model multiplicity, as highlighted by
\ \cite{black_less_2023}, emphasizes that among equally accurate models, one should select those that are less discriminatory.
Additional perspectives include
de-biasing~\cite{bolukbasi_man_2016}, corrective transformations~\cite{dwork2023real}, and work that seeks to move beyond explicit definitions of fairness altogether \cite{jung_algorithmic_2020}.

\subsection{Endogenous Biases: Empirical Evidence}

A substantial body of empirical evidence shows that individuals’ self-presentation styles introduce systematic disparities in evaluation outcomes.
For example, East Asian employees tend to exhibit a modesty bias in self-ratings of job performance, leading to lower self-assessments despite comparable actual performance \cite{farh1991cultural,jiayuanyu_modesty_1993,cho_matter_2023}.
Such disparities highlight how cultural norms of modesty versus self-promotion can shape the information available to evaluators, and thereby affect professional advancement.

Research on self-presentational behavior shows that applicants strategically adjust how they present themselves during hiring processes, often influencing evaluations in ways that are independent of their actual skills \cite{konig_applicants_2011}.
More recently, studies have explored how AI systems interact with such biases, showing both risks and opportunities.
For instance, \cite{hofmann_ai_2024} and \cite{guenzel_ai_2025} analyze the role of AI in recruitment and assessment, noting that algorithmic tools can either amplify existing disparities or help to mitigate them if designed carefully.

Relatedly, \cite{eloundou_first-person_nodate} investigates biases in chatbot responses linked to users’ names, illustrating how seemingly innocuous identity cues can alter model behavior.
\cite{bolukbasi_man_2016} shows that widely used word embeddings encode gender stereotypes, raising concerns that linguistic representations themselves can reproduce and amplify social biases.
Most recently, \cite{agarwal_ai_2024} demonstrates that Western-centric AI suggestions led Indian participants to adopt Western writing styles, homogenizing their work toward Western norms and diminishing cultural nuance.

\subsection{LLMs as Interviewers}
In recent years, AI has increasingly been employed as an interviewer across a range of contexts, including the elicitation of political opinions \citep{wuttke2024ai}, qualitative interviews \citep{chopra2023conducting}, quantitative telephone surveys \citep{leybzon2025ai}, and even in assisting with questionnaire design \citep{adhikari2025exploring}.
\cite{wangadaptive} further proposes an adaptive elicitation framework that leverages LLMs to select questions sequentially so as to reduce uncertainty about latent traits.
While these approaches emphasize scalability, efficiency, or data quality, our work differs in explicitly addressing fairness by ensuring that elicited skill assessments are not systematically biased by individual differences in self-presentation style.

\section{Methods}\label{sec:methods}
\subsection{Problem Setup}

We study the \emph{skill elicitation problem}: given an individual $j$, the goal is to determine whether they possess a target skill. The ground-truth label is denoted $y^j \in \{0,1\}$, and our predictor produces an estimate $\hat y^j \in [0,1]$.

The elicitation is interactive. At each round $i$, the system asks a question $q_i$ from a fixed bank $\mathcal Q$ (appendix~\ref{app:question_bank}), and the individual responds in free text $s_i$. The dialogue history up to round $i$ is called the \emph{transcript}:
\[
x_i^j =  (q_1^j, s_1^j, \ldots, q_i^j, s_i^j).
\]

Let $\mathcal X$ denote the space of transcripts. Our system learns two components:
1) A \textbf{question selector} $q_\psi : \mathcal X \to \Delta(\mathcal Q)$, parameterized by $\psi$, which maps the transcript $x_i^j$ to a probability distribution over the remaining questions.
2) A \textbf{score function} $f_\phi : \mathcal X \to [0,1]$, parameterized by $\phi$, which outputs the predicted probability $\hat y^j$ of skill possession.

Each response $s_i$ is embedded using a pretrained text encoder $\operatorname{BERT}()$, concatenated with its question index, and aggregated with a GPT-style attention mechanism $\operatorname{ATTN}()$. For example:
\begin{align*}
q_\psi(x_i) &= \operatorname{ATTN}_\psi(q_1 \circ \operatorname{BERT}(s_1), \dots, q_i \circ \operatorname{BERT}(s_i)), \\
f_\phi(x_i) &= \operatorname{ATTN}_\phi(q_1 \circ \operatorname{BERT}(s_1), \dots, q_i \circ \operatorname{BERT}(s_i)),
\end{align*}
where $\circ$ denotes concatenation.

\subsection{Optimization Objectives}\label{sec:opt}

Our learning problem can be cast in an actor--critic framework~\citep{sutton2018reinforcement}, where the question selector $q_\psi$ plays the role of the \emph{actor} and the score function $f_\phi$ serves as the \emph{critic}. The actor is responsible for deciding which question to ask next, while the critic evaluates the quality of the interaction by predicting whether the individual possesses the target skill. The two models are updated jointly and influence one another during training. Below we detail the objectives for each component.

\paragraph{Certainty-maximizing policy learning.}
A na\"{i}ve approach would be to update the question selector using the final prediction error $|y^j - \hat y^j|$ as the reward. However, this leads to very sparse and delayed feedback: a sequence of questions may only be judged good or bad once the entire interview is complete. To provide a more informative training signal, we instead optimize the question selection policy for immediate \emph{information gain} \citep{wangadaptive}.

For a transcript $x_i^j$ and candidate next question $q_{i+1}^j$, we simulate $M$ steps into the future using the response simulator $s()$. This is a generative model of an individual, provided by our trained language model; see Section~\ref{sec:LM fine}. We use $x_{i+M}^j$ to compute the rollout-based {\em certainty} measure
\[
Z({ q^j_{i+1}}) = 2\big|f_\phi(x_{i+M}^j) - 0.5\big|.
\]
Intuitively, $Z({ q^j_{i+1}})$ is large when asking ${ q^j_{i+1}}$ leads to states in which the score function $f_\phi$ makes more confident predictions. In other words, the policy is rewarded for selecting questions that are expected to most reduce predictive uncertainty, forcing prediction errors to be quickly corrected.
Algorithm~\ref{alg:z-new} outlines this rollout-based calculation for ${ q^j_{i+1}}$: it samples a length $M$ extension of the current transcript, for some parameter $M$, beginning with ${ q^j_{i+1}}$ and running for $M-1$ more steps, and calculates the certainty of $q^j_{i+M}$.
  We call this the {\em rollout-based certainty (for ${ q^j_{i+1}}$).}
\begin{algorithm}[t]
\caption{Rollout-Based Certainty $Z$}
\label{alg:z-new}
\begin{algorithmic}
\STATE{\bf Input:} Transcript $x_i^j$, candidate question $q$, depth $M$, models $f_\phi$, $q_\psi$, simulator $s()$
\STATE{\bf Output:} $Z$
\STATE Create $x_{i+1}^j$ by appending $q$ and its simulated response
\FOR{$m=1$ to $M-1$}
    \STATE Sample next question $q'$ from $q_\psi(x_{i+m}^j)$
    \STATE Append $q'$ and simulated response to form $x_{i+m+1}^j$
\ENDFOR
\STATE Compute $Z = 2|f_\phi(x_{i+M}^j) - 0.5|$
\STATE \textbf{return} $Z$
\end{algorithmic}
\end{algorithm}

Given this reward definition, we optimize the question selector using the REINFORCE algorithm. Specifically, the parameters $\psi$ are updated by~\footnote{We use the log-derivative trick following the Proximal Policy Optimization (PPO) algorithm~\cite{schulman2017proximal,weng2018rl}:
$\nabla_\psi \E_{p(q\mid\psi)}[Z(q)]=\E_{p(q\mid\psi)}[\nabla_\psi \log p(q\mid\psi)\cdot Z(q)]$}
\[
\psi \leftarrow \psi + \alpha \gamma Z({  q^j_{i+1}}) \nabla_\psi \log q_\psi(q_{i+1}^j \mid x_i^j),
\]
where $\alpha$ is the learning rate, $\gamma$ is a discount factor (both set to $1$ in our case), and the notation $q_\psi(q_{i+1}^j \mid x_i^j)$ is used to denote the probability mass at $q_{i+1}^j$ in the probability distribution $q_\psi(x_i^j)$. This update increases the probability of choosing questions that are expected to maximize certainty gain. This update encourages $q_\psi$ to increase the probability of selecting informative questions.

\paragraph{Supervised training of the score function.}
The critic $f_\phi$ is trained to approximate the ground-truth label $y^j$. Unlike the policy, the critic has access to the true skill label during training and can therefore be updated with a direct supervised loss. Specifically, we use mean squared error (MSE):
\[
\mathcal{L}_\text{MSE}(\phi) = \frac{1}{B} \sum_{j=1}^B \big(f_\phi(x_i^j) - y^j\big)^2.
\]
The gradient update for $\phi$ is
\[
\phi \leftarrow \phi - \beta (f_\phi(x_i^j) - y^j)\nabla_\phi f_\phi(x_i^j),
\]
with learning rate $\beta$. This ensures that $f_\phi$ gradually improves its skill predictions as more rich transcripts are observed.

The objectives of $q_\psi$ and $f_\phi$ are interdependent. As the critic improves, its uncertainty estimates $Z$ become sharper, which provides the actor with a more reliable reward signal. Conversely, as the actor learns to ask more informative questions, the critic observes richer transcripts and thus achieves higher predictive accuracy. This coupling drives a virtuous cycle: the actor guides exploration, while the critic supplies feedback, leading to a jointly optimized interactive elicitation process.

\subsection{Model Fairness}\label{sec:fair}

While the optimization objectives above reduce endogenous (self-reporting) bias and ensure that the score function $f_\phi$ becomes accurate on average, they do not guarantee that the learned model is free of bias. In particular, even after successful elicitation, skill evaluation may correlate with personality traits or communication styles that are irrelevant to the actual skill. If left unchecked, the model could systematically overestimate or underestimate individuals from certain personality groups, leading to biased outcomes.

\paragraph{Multi-Accuracy.}
To mitigate this, we require the predictor to be \emph{multi-accurate} with respect to a family of auxiliary functions. Let $\mathcal C$ denote this family\footnote{In our proof of concept

$|\mathcal C|=6$.}. Each $c \in \mathcal C$ maps a transcript $x_i^j$ to a real-valued score in $[0,1]$, representing, for instance, how strongly the individual exhibits a certain trait.

A predictor $f_\phi$ is said to be $(\mathcal C, \epsilon)$-multi-accurate if
\begin{align}\label{eq:multi_acc}
\sup_{c \in \mathcal C} \Big| \E_j \big[c(x_i^j)\cdot(f_\phi(x_i^j) - y^j)\big] \Big| \le \epsilon.
\end{align}
This condition ensures that the prediction error of the score function is approximately uncorrelated with all functions $c \in \mathcal C$: no personality dimension (or their linear combinations) can explain away the residual bias.

\paragraph{Enforcing Multi-Accuracy.}
To enforce multi-accuracy, we adapt the procedure of \citep{gopalan2023loss}. The idea is to correct the raw predictor $f_\phi$ by adding a linear combination of the auxiliary functions, followed by a transfer function. Concretely, we define
\[
f_\phi^*(x) = \sigma\!\left(\sum_{c\in \mathcal C} l_c \, c(x) + \sigma^{-1}(f_\phi(x))\right),
\]
where $\sigma(t)=1/(1+\exp(-t))$ is the sigmoid function, $\sigma^{-1}$ is its logit, and $\{l_c\}$ are calibration weights. Intuitively, the correction shifts the logit of $f_\phi(x)$ depending on the personality scores $c(x)$ so that the residual correlation between errors and traits is minimized. The weights $\{l_c\}$ are obtained by solving a convex optimization problem derived from Theorem~5.6 of \citep{gopalan2023loss} (see the theorem statement in Appendix~\ref{sec:supp-pf}):
\begin{align}\label{loss:ma}
    \mathcal{L}_{g}(y,z) &= -\bigl(y \log (z) + (1 - y) \log (1 - z)\bigr)\\
    &=-\frac{1}{N} \sum_{j=1}^N \Biggl[y^j \log \Biggl(\sigma\Bigl(\sum_{c\in\mathcal C} l_c c(x_i^j) + \sigma^{-1}(f(x_i^j))\Bigr)\Bigr) \notag\\
    &\quad\quad\quad\quad\quad\quad\quad + (1 - y^j) \log \Bigl(1 - \sigma\Bigl(\sum_{c\in\mathcal C} l_c c(x_i^j) + \sigma^{-1}(f(x_i^j))\Bigr)\Biggr) \Biggr]\notag\\
    &\quad+ \epsilon\cdot \sum_{c\in\mathcal C}|l_c| \notag
\end{align}
This yields a predictor $f_\phi^*$ that is provably multiaccurate with respect to $\mathcal C$.

\paragraph{Integration with training.}

Fairness calibration is interleaved with the actor-critic updates described earlier. After a number of standard training steps, we perform a calibration phase: estimate $\{l_c\}$ on a batch of transcripts, update the calibrated predictor $f_\phi^*$, and then use $f_\phi^*$ as $f_\phi$ for the next batch. This alternating procedure ensures that improvements in predictive accuracy are continually sharpened by multiaccuracy post-processing\footnote{The imposition of multiaccuracy \textit{improves} predictive accuracy: since $f_\phi^*$ is the optimal solution to the optimization problem, $f_\phi^*$ has smaller or equal cross entropy loss than $f_\phi$. This implies that the KL divergence with the "ground truth" $p^*$ is shrinking: $KL(p^* || f_\phi^*) \le KL(p^* || f_\phi)$}. If the calibration condition is ever violated (i.e., the bound above exceeds $\epsilon$, say, due to a failure of the optimizer), training reverts to the last feasible state.

\subsubsection{Generalization Threshold}\label{sec:threshold}

The number of samples needed to guarantee fairness generalization can be calculated based on the value of $\epsilon$ by

\begin{align}\label{eq:generalization}
O(\frac{1}{\epsilon^2} \log \frac{|C|}{\delta}).
\end{align}
where $\delta$ is the failure rate. In our instantiation we have $\delta =0.000001$ and $|C|=6$.

\subsection{Training Algorithm}

Algorithm~\ref{alg:main} summarizes the full training loop. Each iteration alternates between:
(i) sampling transcripts,
(ii) computing rollout-based uncertainty values $Z$,
(iii) updating $q_\psi$ and $f_\phi$ with stochastic gradients, and
(iv) periodically recalibrating $f_\phi$ to enforce multi-accuracy.
\begin{algorithm}[t]
\caption{Rollouts Algorithm}\label{alg:rollouts}
\begin{algorithmic}
\STATE{\bf Inputs:} $x^j_i$, $N$, $M$, $f_\phi()$ $q_\psi()$, $s()$
\STATE{\bf Output:} $Z^j_i$, $q^j_i$
\STATE{\bf initialize} $Z^j_i = 0$, $q^j_{i+1} = 0$
\STATE{ Assume for simplicity: $N \le |Q|$}
\STATE{Calculate ordering of next questions $q_1, q_2, \dots, q_N$ using $q_\psi(x^j_i)$ by taking the top $N$}
\FOR{$n = 1$ to $N$}
    \STATE{Create state $x^j_{i+1}$ after $q_n$ using $s()$}
    \FOR{$m=1$ to $M-1$}
        \STATE{Select next question $q$ using $q_\psi (x^{j,n}_{i+m})$}
        \STATE{Create state $x^{j,n}_{i+m}$ by adding $q$ to $x^{j,n}_{i+m}$ then using $s()$}
    \ENDFOR
    \STATE{Calculate $Z^{j,n}_{i+M} = 2|f_\phi(x^{j,n}_{i+M}) - .5|$}
    \IF{$Z^j_{i,n} > Z^j_i$}
        \STATE{$Z^j_i = Z^j_{i,n}$}
        \STATE{$q^j_{i+1} = q_n$}
    \ENDIF
\ENDFOR
\STATE{\bf return} $Z^j_i$, $q^j_{i+1}$
\end{algorithmic}
\end{algorithm}

\begin{algorithm}[ht]
\caption{Equitable Elicitation Algorithm - General Solution}\label{alg:main}
\begin{algorithmic}
\STATE{\bf initialize Fixed Values:} $\mathcal{Q}$, $\mathcal{C}$, $s()$
\STATE{\bf initialize Learned Variables:} $\psi\sim \mathcal{U}$, $\phi\sim \mathcal{U}$, $l_c=0$, $Z=0$
\STATE{\bf initialize Hyper-parameters:} $\alpha$, $\beta$, $\epsilon$, $\gamma$, $N$, $B$, $i_{\text{max}}$
\WHILE{$\phi$ and $\psi$ not converged}
    \FOR {$b^\text{fair}=1$ to $B_\text{fair}$}
        \STATE{Initialize empty batches $\mathcal{B}_x = \{\}$, $\mathcal{B}_Z = \{\}$, $\mathcal{B}_y = \{\}$, $\mathcal{B}_q = \{\}$}
        \STATE{If not first run}
        \STATE{Initialize batches from $\mathcal{B}_x^\text{fair}$}
        \STATE{Pick random $i< i_{\text{max}}$}
        \STATE{$i_{batch} = i$}
        \FOR{$b = 1$ to $B$}
            \STATE{Sample $j$ from $s()$}
            \STATE{Form $x^j_i$ based on $s()$ with  $q_\psi()$}
            \STATE{Calculate $Z^j_i$, $q^j_{i+1}$ using Algorithm~\ref{alg:rollouts} with $x^j_i$, $N$, $f_\phi()$, and $q_\psi()$}
            \STATE{Append $x^j_i$ to $\mathcal{B}_x$, $Z^j_i$ to $\mathcal{B}_Z$, $y^j$ to $\mathcal{B}_y$, and $q^j_{i+1}$ to $\mathcal{B}_q$}
        \ENDFOR
        \STATE{Update $\psi$ using batch $\mathcal{B}_Z$ and $\mathcal{B}_x$ via Adam}
        \STATE{Update $\phi$ using batch $\mathcal{B}_y$, $\mathcal{B}_q$, and $\mathcal{B}_x$ via Adam}

        \ENDFOR
        \STATE{Initialize batches $\mathcal{B}^\text{fair}_x = \{\}$, $\mathcal{B}^\text{fair}_y = \{\}$}
        \STATE{Calculate generalization threshold $\tau$, using Equation~\ref{eq:generalization}}
        \WHILE{$|\mathcal{B}^\text{fair}_x|<\tau$ and $|\mathcal{B}^ \text{fair}_y|<\tau$}
            \STATE{Pick random $i< i_{\text{max}}$}
            \STATE{Add sample to $\mathcal{B}^\text{fair}_x$ and $\mathcal{B}^\text{fair}_y$, as above}
        \ENDWHILE
        \STATE{Calculate $l_c$ for each $c\in\mathcal{C}$ using $\mathcal{B}^\text{fair}_x$ and $\mathcal{B}^\text{fair}_y$ via linear optimizer}
        \STATE{Create $f_\phi^*$ with $l_c$}
        \STATE{Generate test samples $\mathcal{B}^\text{test}_x$, $\mathcal{B}^\text{test}_y$ of size $b^\text{fair}$}
        \STATE{Verify $f_\phi^*$ satisfies Equation~\ref{eq:multi_acc} using $\mathcal{B}^\text{test}_x$, $\mathcal{B}^\text{test}_y$ and $l_c$ }
        \IF{not satisfied}

            \STATE{\textbf{Break}, failure restart at previous good state}
        \ENDIF

\ENDWHILE
\end{algorithmic}
\end{algorithm}

\subsection{Theoretical Guarantee}
We have implicitly assumed that our hypothesis class is sufficiently rich to capture the output of the multi-accuracy algorithm.  Formally, we have the following result, whose proof appears in Appendix~\ref{sec:supp-pf}.

\begin{theorem}\label{thm:ma}
Let $\mathcal F=\{f_\phi: \phi \in \Phi\}$ be the hypothesis class, and suppose the solution to the loss function \eqref{loss:ma}, $f^*$, lies within $\epsilon^*$ of $\mathcal F$. Then $f^*$ is $(\mathcal C, \epsilon+\epsilon^*)$-multi-accurate at every iteration of the interactive procedure. \qed
\end{theorem}

\clearpage
\section{A Specific Instantiation}\label{sec:results}

We have created a prototype {\em elicitation model} that interviews (synthetic) people of differing presentation style to elicit information about their skills.  The particular setting we have in mind is joining a professional networking platform, at which time users provide employment-relevant information.  The goal is to mitigate self-presentation differences and capture relevant facts, creating for the platform an internal representation of skills\footnote{The question of whether the platform shares this with the user is orthogonal to our work.}.

\begin{figure}
    \centering
    \includegraphics[width=1\linewidth]{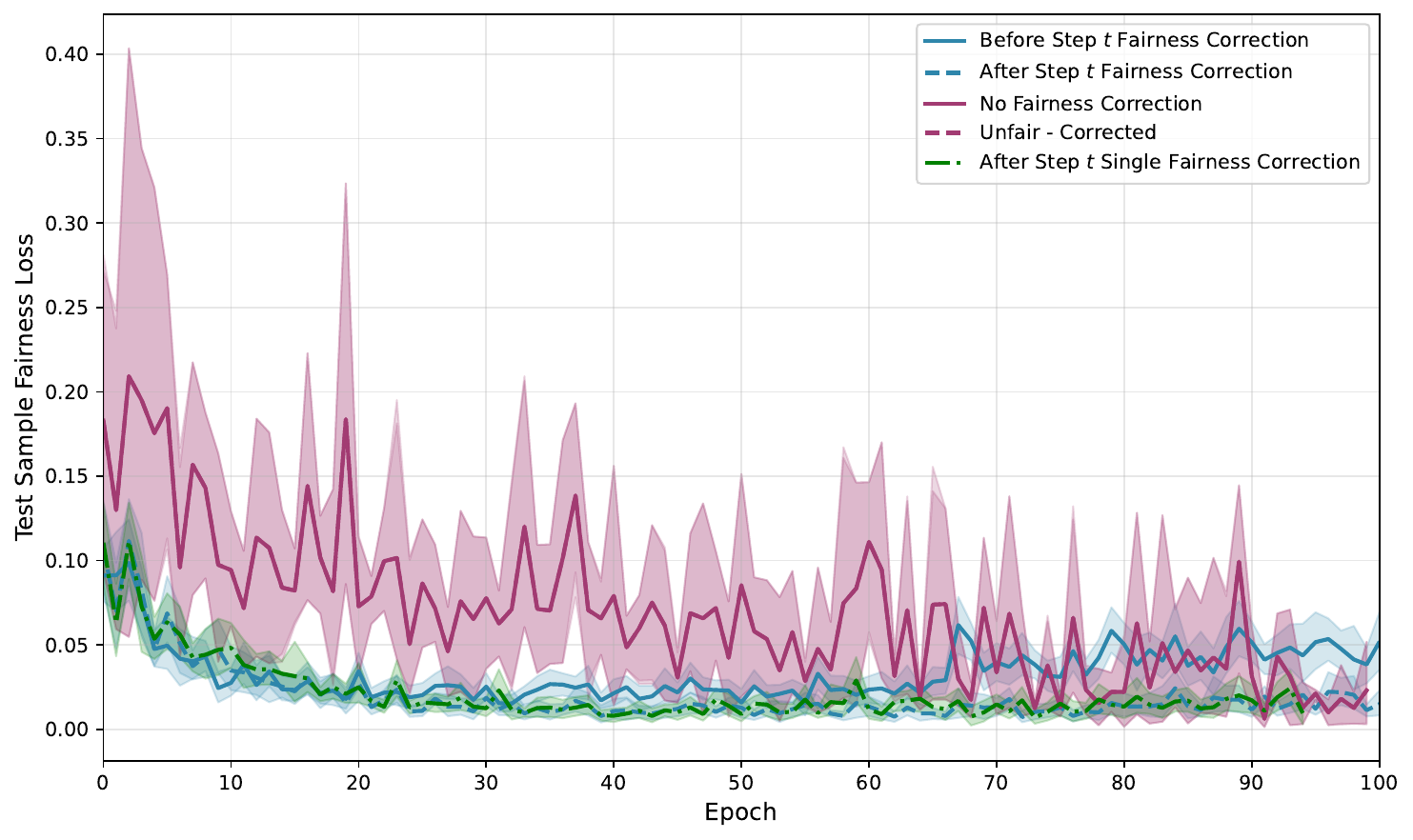}
    \caption{Fairness loss (\textit{Equation}~\ref{loss:ma}) on test samples during training with the fairness correction ({\color{blue}blue}), with the fairness correction applied at all previous epochs ({\color{blue}blue dashed}), and with the fairness correction never applied ({\color{red}red}). Error bars are bootstrap confidence intervals.}
    \label{fig:loss}
\end{figure}

We evaluate the prototype on users synthesized with our trained language model, measuring both accuracy loss ($\Sigma_j |{\rm round}(f_\phi(x^{j}_{i}))-y^j|$, and the formal quantification of multiaccuracy loss from Equation~\ref{loss:ma} (see Figures~\ref{fig:acc} and~\ref{fig:loss}).
To our knowledge, no previous work in reinforcement learning measures, let alone guarantees, multi-accuracy as a core requirement of the training.

Our procedure is inspired by standard interview practices, to wit we frame the interview as a sequence of questions, where each question is chosen from a fixed set of possible questions based on the current state of the interview (see Appendix~\ref{app:question_bank}). The user responds to each question in turn, and at the end of the conversation the model attempts to determine if the user possesses a specific skill.

This work is motivated by the empirically observed endogenous biases in self-description during on-boarding on hiring platforms\citep{Lindstrom2017womensEquality,LinkedIn2017,murciano2022missing} in technical fields. We focus on a component of this problem, personality based biases. We attempt to create a structured conversation process that (1) obtains the data necessary for skills determination and (2) is fair with respect to the \textit{Big 5} personality traits (Openness, Conscientiousness, Extroversion, Agreeableness, Neuroticism).

\subsection{Model Training Procedures}

\subsubsection{Language Model Fine-Tuning}
\label{sec:LM fine}

We start with a set of sample profiles collected from public profiles on a large professional networking site. These profiles were then filtered to remove entries that were missing fields or in which the summary is too short (fewer than 50 characters).
The remaining profiles were then split into training, validation, and test sets.

The profiles were converted into a structured text format, with a fixed set of key-value pairs, with each key representing a field in the profile (e.g., \texttt{"full\_name"}, \texttt{"gender"}, \texttt{"industry"}, \texttt{"skills"}, \texttt{"summary"}, etc.). This structured text format was then used to fine tune \texttt{Meta-Llama-3.1-8B}, creating a model that generates profile summaries based on the other fields in the profile. This is not a general purpose language model, but rather a method of sampling from the space of valid profile summaries; nothing is done to remove the biases inherent in the dataset at this stage.
We use this language model to "respond" to questions by generating counterfactual responses to questions, see Appendix~\ref{app:question_bank} for the list of possible counterfactuals.

\subsubsection{RL Model Training Procedure \texorpdfstring{($f$ and $q$)}{(f and q)}}

\xhdr{Conversation Format} Our structured conversations are a series of counterfactual questions~\footnote{This design choice too is expedient, as these are compatible with our particular language model and the data on which it is trained.}. We model truthful users who manifest no intended deception.  We start with a base profile that represents the user's ground truth.  We then ask questions by appending skills to the profile, having the language respond with a summary that incorporates these skills. Thus we can construct a series of self-summaries that represent the user under different (self-assigned) skill sets.  The training objective is to determine if the user possesses a specific skill based on their sequence of summaries. Note that this creates the need for the two decision making processes as outlined in Algorithm~\ref{alg:main}, with $q_\psi(x_i)$ (the question model) selecting which sequence of self-summaries $f_\phi(x_i)$ (the score function) will encounter.

Our structured conversation format allows us to frame the process as the reinforcement learning problem described in section~\ref{sec:opt}. Given a base profile, it either does or does not possess a target skill. Our end goal is to create a fair scoring function ($f_\phi(x_i)$) that takes the sequence of self summaries and predicts whether or not the interviewee has the target skill.

Given $f_\phi$ we can then assign the correct answer to have a reward of +1, allowing us to attempt to learn a policy that maximizes the accuracy of the final answer. We use a simple policy gradient method, \textit{REINFORCE}~\citep{sutton1999policy}, to learn a policy ($q_\psi$) that selects the next skill, but use reducing uncertainty as the final reward instead of accuracy (as in~\citep{wangadaptive,cai2024active,blei2003latent}). These two functions combine to create an actor-critic model~\citep{konda1999actor}, where the actor is $q_\psi$ and the critic is $f_\phi$.

\subsection{Results}

\begin{figure}
    \centering
    \includegraphics[width=1\linewidth]{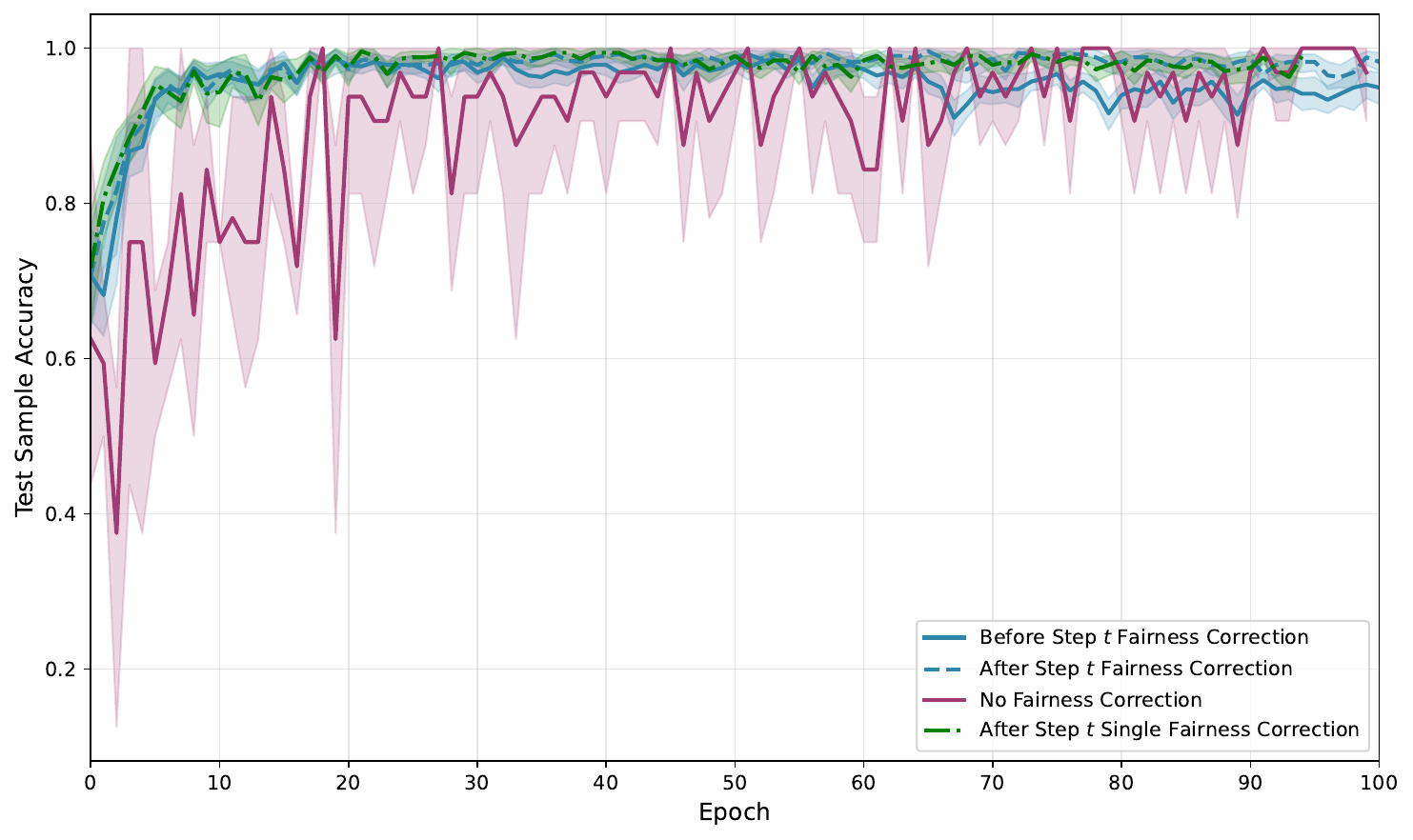}
    \caption{Accuracy on test samples during training with fairness correction at all epochs ({\color{blue}blue}), with the fairness correction applied at all previous epochs ({\color{blue}blue dashed}), with the fairness correction never applied ({\color{red}red}), and with the fairness correction applied only at step~$t$ ({\color{green} green}). Error bars are bootstrap confidence intervals.}
    \label{fig:acc}
\end{figure}

We train both $f_{\phi}$ and $q_\psi$ using

a small (400k parameter) transformer based neural network,

with the transcripts being preprocessed via an embedding model\footnote{We used \hyperlink{https://huggingface.co/sentence-transformers/all-MiniLM-L6-v2}{\texttt{sentence-transformers/all-MiniLM-L6-v2}}~\citep{reimers-2019-sentence-bert}, which is based on \cite{wang2020minilm}}. $f_\phi$ and $q_\psi$ do not share parameters, but have identical designs modulo the final layer. We used the language model to simulate the platform users, with half being initialized with real profiles from our validation set, and half being fully synthetic. The skill in question is leadership, with 20\% of users having it during our simulations. We allowed the model to ask 2 questions after the initial self description, and our rollouts were for an additional 2 questions.
As described above, questions are posed as new skills the user may have.
Each step was a batch of 8 structured conversations, with the fairness calibration being done every 32 conversations. We ran the process for 100 epochs  as each of the losses ({\it i.e.}, on $f_\phi()$ and $q_\psi$ respectively) converged by that point (see Figure~\ref{fig:acc}).

When trained with the above procedure, the fairness loss was 0.04803 before the correction, which is above our target threshold of .01, after the multi-accuracy correction it was 0.00223, this procedure also increases the accuracy from .937, to 1.0 on the test set, \textit{i.e.,} samples that were not used in training. Note that we display samples that show variance after the model is fully trained (around step 60 on this non-cherry-picked run), this phenomena occurs on some training runs likely due to overfitting.

Multi-accuracy improves overall accuracy.  However,
applying the multi-accuracy correction only at step $t$ yields accuracy comparable to that obtained by applying the correction repeatedly up to step~$t$
(see Figure~\ref{fig:acc}).  Similarly, applying multi-accuracy repeatedly up to step $t$ increases confidence at $t$, but only a bit more than applying it just once at time $t$ (see Figure~\ref{fig:Z-scores} in Appendix~\ref{app:Z-scores}).

\section{Discussion}\label{sec:discussion}
In this work we address two sources of bias: endogenous biases arising in self-reporting, and systematic model biases arising during training.
The former is accomplished via an AI that interacts with the users and permits them to respond in their own voice, the latter is achieved by imposing

post-hoc theoretical guarantees of fairness.
Our work presents an existence proof that fairness can be improved, beyond simply ensuring multi-accuracy, in an interactive social setting.

Our key contributions are:
\begin{enumerate}
\item In {\em computational social science} we provide a novel example of the use of LLMs, finetuning an existing language model to create synthetic humans on which we train an equitable and fair scoring function.

\item We fine-tune an existing LLM to obtain a large language model that can accurately reproduce a complex dataset and allows us to test our methods without risking harm to individuals.\footnote{We started from a public dataset. Starting from confidential data would require privacy protections, for example, via differential privacy.} That said, we are not advocating for replacing human subjects research with LLMs, and as we noted in Section~\ref{sec:results}, these methods should be used in addition to real human interviewers' analysis.
\item We address endogenous biases in individuals' self-reports {\em by eliciting additional information}.
\item Our technique permits individuals to {\em speak in their own voice}.
\item We incorporate model bias correction, specifically, multi-accuracy with respect to personality types, at regular intervals during training, improving accuracy over what is obtained with no model bias correction.

\end{enumerate}

To our knowledge, all prior work on algorithmic fairness "makes do" with whatever data are available. Ours is the first system to actively elicit additional information in order to address feature-insufficiency for fairness.

\xhdr{Future Work}
While we focus on the interview context with a single skill, equitable elicitation can be applied to estimating multiple skills simultaneously.  Moreover, chat bots are growing in usage~\citep{stohr2024perceptions}, and equitable elicitation would be beneficial in many other realms, such as making medical or financial recommendations, which often rely on analysis of transcripts.

Looking through the lens of theory, how should we formalize the goal of equitable elicitation?  In our setting, we were able to assume there is a ground truth that can be discovered through interaction with the user; there is no aleatoric uncertainty. In such a setting we might aim for a condition analogous to standard multi-group fairness conditions, eg, that the absolute or squared error have low covariance with the distinguishing functions $c\in C$.

Although we expect our approach also to be applicable for improving predictors of {\em future} events, for example, whether a candidate will be promoted within 2 years of hiring, the definition of equitable elicitation in this case now would have to handle differences in inherent uncertainty that truly are related to the functions in $C$ -- even if the  uncertainty does not inhere in the individual but rather in the interaction between the individual and the work environment.

 Finally, our work was inspired by research showing that endogenous biases in self-presentation are gendered.  Looking more broadly, how can we design algorithms to detect when collected features are insufficient for fairness?

\bibliography{99-bib}

@article{murciano2022missing,
  title={Missing women in tech: The labor market for highly skilled software engineers},
  author={Murciano-Goroff, Raviv},
  journal={Management Science},
  volume={68},
  number={5},
  pages={3262--3281},
  year={2022},
  publisher={INFORMS}
}

@inproceedings{mukherjee2020two,
  title={Two simple ways to learn individual fairness metrics from data},
  author={Mukherjee, Debarghya and Yurochkin, Mikhail and Banerjee, Moulinath and Sun, Yuekai},
  booktitle={International conference on machine learning},
  pages={7097--7107},
  year={2020},
  organization={PMLR}
}

@inproceedings{kleinberg2018algorithmic,
  title={Algorithmic fairness},
  author={Kleinberg, Jon and Ludwig, Jens and Mullainathan, Sendhil and Rambachan, Ashesh},
  booktitle={Aea papers and proceedings},
  volume={108},
  pages={22--27},
  year={2018},
  organization={American Economic Association 2014 Broadway, Suite 305, Nashville, TN 37203}
}

@inproceedings{jung2021moment,
  title={Moment multicalibration for uncertainty estimation},
  author={Jung, Christopher and Lee, Changhwa and Pai, Mallesh and Roth, Aaron and Vohra, Rakesh},
  booktitle={Conference on Learning Theory},
  pages={2634--2678},
  year={2021},
  organization={PMLR}
}

@misc{weng2018rl,
  author       = {Lilian Weng},
  title        = {A (Long) Peek into Reinforcement Learning},
  year         = {2018},
  month        = feb,
  url          = {https://lilianweng.github.io/posts/2018-02-19-rl-overview/},
  note         = {Accessed: December 14, 2025}
}

@article{schulman2017proximal,
  title={Proximal policy optimization algorithms},
  author={Schulman, John and Wolski, Filip and Dhariwal, Prafulla and Radford, Alec and Klimov, Oleg},
  journal={arXiv preprint arXiv:1707.06347},
  year={2017}
}

@article{chouldechova2017fair,
  title={Fair prediction with disparate impact: A study of bias in recidivism prediction instruments},
  author={Chouldechova, Alexandra},
  journal={Big data},
  volume={5},
  number={2},
  pages={153--163},
  year={2017},
  publisher={Mary Ann Liebert, Inc. 140 Huguenot Street, 3rd Floor New Rochelle, NY 10801 USA}
}

@book{sutton2018reinforcement,
  title     = {Reinforcement learning: An introduction},
  author    = {Sutton, Richard S and Barto, Andrew G},
  year      = {2018},
  publisher = {MIT press}
}

@inproceedings{dwork2023real,
  title={From the real towards the ideal: Risk prediction in a better world},
  author={Dwork, Cynthia and Reingold, Omer and Rothblum, Guy N},
  booktitle={4th Symposium on Foundations of Responsible Computing (FORC 2023)},
  pages={1--1},
  year={2023},
  organization={Schloss Dagstuhl--Leibniz-Zentrum f{\"u}r Informatik}
}

@article{ilvento2019metric,
  title={Metric learning for individual fairness},
  author={Ilvento, Christina},
  journal={arXiv preprint arXiv:1906.00250},
  year={2019}
}

@inproceedings{kearns2018preventing,
  title={Preventing fairness gerrymandering: Auditing and learning for subgroup fairness},
  author={Kearns, Michael and Neel, Seth and Roth, Aaron and Wu, Zhiwei Steven},
  booktitle={International conference on machine learning},
  pages={2564--2572},
  year={2018},
  organization={PMLR}
}

@article{mohr2014women,
  title={Why women don’t apply for jobs unless they’re 100\% qualified’Harvard Business Review},
  author={Mohr, TS},
  journal={August 25th},
  year={2014}
}

@article{ahuja2024right,
  title={Right byte or left out? Gender differences in self-presentation among job-seekers on LinkedIn in India},
  author={Ahuja, Kanika K},
  journal={Discover Psychology},
  volume={4},
  number={1},
  pages={59},
  year={2024},
  publisher={Springer}
}

@article{deschacht2017cross,
  title={Cross-cultural differences in self-promotion: A study of self-citations in management journals},
  author={Deschacht, Nick and Maes, Birgitt},
  journal={Journal of Occupational and Organizational Psychology},
  volume={90},
  number={1},
  pages={77--94},
  year={2017},
  publisher={Wiley Online Library}
}

@article{LinkedIn2017,
title = {Women’s Equality Day: A Look At Women in The Workplace in 2017},
author = {Rachel Bowley Lindstr\"{o}m},
journal = {LinkedIn Official Blog},
note = {\url{https://www.linkedin.com/blog/member/career/womens-equality-day-a-look-at-women-in-the-workplace-in-2017}},
year = {2017}
}

@article{gopalan2022loss,
  title={Loss minimization through the lens of outcome indistinguishability},
  author={Gopalan, Parikshit and Hu, Lunjia and Kim, Michael P and Reingold, Omer and Wieder, Udi},
  journal={arXiv preprint arXiv:2210.08649},
  year={2022}
}

@article{gopalan2021omnipredictors,
  title={Omnipredictors},
  author={Gopalan, Parikshit and Kalai, Adam Tauman and Reingold, Omer and Sharan, Vatsal and Wieder, Udi},
  journal={arXiv preprint arXiv:2109.05389},
  year={2021}
}

@inproceedings{hebert2018multicalibration,
  title={Multicalibration: Calibration for the (computationally-identifiable) masses},
  author={H{\'e}bert-Johnson, Ursula and Kim, Michael and Reingold, Omer and Rothblum, Guy},
  booktitle={International Conference on Machine Learning},
  pages={1939--1948},
  year={2018},
  organization={PMLR}
}

@inproceedings{dwork2021outcome,
  title={Outcome indistinguishability},
  author={Dwork, Cynthia and Kim, Michael P and Reingold, Omer and Rothblum, Guy N and Yona, Gal},
  booktitle={Proceedings of the 53rd Annual ACM SIGACT Symposium on Theory of Computing},
  pages={1095--1108},
  year={2021}
}

@article{sutton1999policy,
  title={Policy gradient methods for reinforcement learning with function approximation},
  author={Sutton, Richard S and McAllester, David and Singh, Satinder and Mansour, Yishay},
  journal={Advances in neural information processing systems},
  volume={12},
  year={1999}
}

@article{blei2003latent,
  title={Latent dirichlet allocation},
  author={Blei, David M and Ng, Andrew Y and Jordan, Michael I},
  journal={Journal of machine Learning research},
  volume={3},
  number={Jan},
  pages={993--1022},
  year={2003}
}

@article{cai2024active,
  title={Active Exploration via Autoregressive Generation of Missing Data},
  author={Cai, Tiffany Tianhui and Namkoong, Hongseok and Russo, Daniel and Zhang, Kelly W},
  journal={arXiv preprint arXiv:2405.19466},
  year={2024}
}

@misc{Lindstrom2017womensEquality,
  author       = {Rachel Bowley Lindstr{\"o}m},
  title        = {Women’s Equality Day: A Look At Women in The Workplace in 2017},
  year         = {2017},
  month        = {8},
  day          = {28},
  url          = {https://www.linkedin.com/blog/member/career/womens-equality-day-a-look-at-women-in-the-workplace-in-2017},
  urldate      = {2025-12-05},
  organization = {LinkedIn},
  note         = {LinkedIn Official Blog}
}

@inproceedings{gopalan2023loss,
  title={Loss Minimization Through the Lens Of Outcome Indistinguishability},
  author={Gopalan, Parikshit and Hu, Lunjia and Kim, Michael P and Reingold, Omer and Wieder, Udi},
  booktitle={14th Innovations in Theoretical Computer Science Conference (ITCS 2023)},
  pages={60--1},
  year={2023},
  organization={Schloss Dagstuhl--Leibniz-Zentrum f{\"u}r Informatik}
}

@inproceedings{wangadaptive,
  title={Adaptive Elicitation of Latent Information Using Natural Language},
  author={Wang, Jimmy and Zollo, Thomas P and Zemel, Richard and Namkoong, Hongseok},
  booktitle={ICLR Workshop: Quantify Uncertainty and Hallucination in Foundation Models: The Next Frontier in Reliable AI},
  year={2025}
}

@misc{bolukbasi_man_2016,
	title = {Man is to Computer Programmer as Woman is to Homemaker? Debiasing Word Embeddings},
year={2016},
	url = {http://arxiv.org/abs/1607.06520},
	doi = {10.48550/arXiv.1607.06520},
	shorttitle = {Man is to Computer Programmer as Woman is to Homemaker?},
	number = {{arXiv}:1607.06520},
	publisher = {{arXiv}},
	author = {Bolukbasi, Tolga and Chang, Kai-Wei and Zou, James and Saligrama, Venkatesh and Kalai, Adam},
	urldate = {2024-01-29},
	date = {2016-07-21},
	eprinttype = {arxiv},
	eprint = {1607.06520 [cs, stat]},
}

@misc{black_less_2023,
	location = {Rochester, {NY}},
	title = {Less Discriminatory Algorithms},
	url = {https://papers.ssrn.com/abstract=4590481},
	doi = {10.2139/ssrn.4590481},
	number = {4590481},
	author = {Black, Emily and Koepke, John Logan and Kim, Pauline and Barocas, Solon and Hsu, Mingwei},
	urldate = {2024-02-09},
	date = {2023-10-02},
	langid = {english},
year={2024}
}

@article{wang_learning_nodate,
  title={Learning interpretable fair representations},
  author={Wang, Tianhao and Bu{\c{c}}inca, Zana and Ma, Zilin},
  journal={arXiv preprint arXiv:2406.16698},
  year={2024}
}

@article{cho_matter_2023,
  title={A matter of when, not whether: A meta-analysis of modesty bias in East Asian self-ratings of job performance.},
  author={Cho, Inchul and Hu, Biyun and Berry, Christopher M},
  journal={Journal of Applied Psychology},
  volume={108},
  number={2},
  pages={291},
  year={2023},
  publisher={American Psychological Association}
}

@article{konig_applicants_2011,
	title={Applicants' Self-presentational Behavior across Cultures: Less self-presentation in S witzerland and I celand than in the U nited S tates},
  author={K{\"o}nig, Cornelius J and Hafsteinsson, Leifur G and Jansen, Anne and Stadelmann, Eveline H},
  journal={International Journal of Selection and Assessment},
  volume={19},
  number={4},
  pages={331--339},
  year={2011},
  publisher={Wiley Online Library}
}

@article{jiayuanyu_modesty_1993,
  title={Modesty bias in self-ratings of performance: A test of the cultural relativity hypothesis},
  author={JIAYUANYU and Murphy, Kevin R},
  journal={Personnel Psychology},
  volume={46},
  number={2},
  pages={357--363},
  year={1993},
  publisher={Wiley Online Library}
}

@article{hofmann_ai_2024,
	title = {{AI} generates covertly racist decisions about people based on their dialect},
year=2024,
	rights = {2024 The Author(s)},
	issn = {1476-4687},
journal={Nature},
	url = {https://www.nature.com/articles/s41586-024-07856-5},
	doi = {10.1038/s41586-024-07856-5},
	pages = {1--8},
	journaltitle = {Nature},
	author = {Hofmann, Valentin and Kalluri, Pratyusha Ria and Jurafsky, Dan and King, Sharese},
	urldate = {2024-09-02},
	date = {2024-08-28},
	langid = {english},
	note = {Publisher: Nature Publishing Group},
}

@misc{agarwal_ai_2024,
	title = {{AI} Suggestions Homogenize Writing Toward Western Styles and Diminish Cultural Nuances},
year= 2024,
	url = {http://arxiv.org/abs/2409.11360},
	doi = {10.48550/arXiv.2409.11360},
	number = {{arXiv}:2409.11360},
	publisher = {{arXiv}},
	author = {Agarwal, Dhruv and Naaman, Mor and Vashistha, Aditya},
	urldate = {2024-10-02},
	date = {2024-09-17},
	eprinttype = {arxiv},
	eprint = {2409.11360 [cs]},
}

@article{eloundou_first-person_nodate,
	title = {First-Person Fairness in Chatbots},
    year=2024,
    journal={ICLR},
	author = {Eloundou, Tyna and Beutel, Alex and Robinson, David G and Gu-Lemberg, Keren and Brakman, Anna-Luisa and Mishkin, Pamela and Shah, Meghan and Heidecke, Johannes and Weng, Lilian and Kalai, Adam Tauman},
	langid = {english},
}

@misc{jung_algorithmic_2020,
	title = {An Algorithmic Framework for Fairness Elicitation},
year=2020,
	url = {http://arxiv.org/abs/1905.10660},
	doi = {10.48550/arXiv.1905.10660},
	number = {{arXiv}:1905.10660},
	publisher = {{arXiv}},
	author = {Jung, Christopher and Kearns, Michael and Neel, Seth and Roth, Aaron and Stapleton, Logan and Wu, Zhiwei Steven},
	urldate = {2025-02-14},
	date = {2020-10-14},
	langid = {english},
	eprinttype = {arxiv},
	eprint = {1905.10660 [cs]},
}

@misc{rothblum_probably_2018,
	title = {Probably Approximately Metric-Fair Learning},
	url = {http://arxiv.org/abs/1803.03242},
	doi = {10.48550/arXiv.1803.03242},
	number = {{arXiv}:1803.03242},
	publisher = {{arXiv}},
	author = {Rothblum, Guy N. and Yona, Gal},
	urldate = {2025-02-14},
	date = {2018-07-01},
	eprinttype = {arxiv},
	eprint = {1803.03242 [cs]},
year=2018
}

@article{stohr2024perceptions,
  title={Perceptions and usage of AI chatbots among students in higher education across genders, academic levels and fields of study},
  author={St{\"o}hr, Christian and Ou, Amy Wanyu and Malmstr{\"o}m, Hans},
  journal={Computers and Education: Artificial Intelligence},
  volume={7},
  pages={100259},
  year={2024},
  publisher={Elsevier}
}

@article{konda1999actor,
  title={Actor-critic algorithms},
  author={Konda, Vijay and Tsitsiklis, John},
  journal={Advances in neural information processing systems},
  volume={12},
  year={1999}
}

@inproceedings{reimers-2019-sentence-bert,
  title = "Sentence-BERT: Sentence Embeddings using Siamese BERT-Networks",
  author = "Reimers, Nils and Gurevych, Iryna",
  booktitle = "Proceedings of the 2019 Conference on Empirical Methods in Natural Language Processing",
  month = "11",
  year = "2019",
  publisher = "Association for Computational Linguistics",
  url = "https://arxiv.org/abs/1908.10084",
}

@misc{wang2020minilm,
    title={MiniLM: Deep Self-Attention Distillation for Task-Agnostic Compression of Pre-Trained Transformers},
    author={Wenhui Wang and Furu Wei and Li Dong and Hangbo Bao and Nan Yang and Ming Zhou},
    year={2020},
    eprint={2002.10957},
    archivePrefix={arXiv},
    primaryClass={cs.CL}
}

@misc{guenzel_ai_2025,
	location = {Rochester, {NY}},
	title = {{AI} Personality Extraction from Faces: Labor Market Implications},
	url = {https://papers.ssrn.com/abstract=5089827},
    year=2025,
	doi = {10.2139/ssrn.5089827},
	shorttitle = {{AI} Personality Extraction from Faces},
	number = {5089827},
	publisher = {Social Science Research Network},
	author = {Guenzel, Marius and Kogan, Shimon and Niessner, Marina and Shue, Kelly},
	urldate = {2025-02-18},
	date = {2025-01-09},
	langid = {english},
}

@inproceedings{dwork2012fairness,
  title={Fairness through awareness},
  author={Dwork, Cynthia and Hardt, Moritz and Pitassi, Toniann and Reingold, Omer and Zemel, Richard},
  booktitle={Proceedings of the 3rd innovations in theoretical computer science conference},
  pages={214--226},
  year={2012}
}

@article{hardt2016equality,
  title={Equality of opportunity in supervised learning},
  author={Hardt, Moritz and Price, Eric and Srebro, Nati},
  journal={Advances in neural information processing systems},
  volume={29},
  year={2016}
}

@inproceedings{zemel2013learning,
  title={Learning fair representations},
  author={Zemel, Rich and Wu, Yu and Swersky, Kevin and Pitassi, Toni and Dwork, Cynthia},
  booktitle={International conference on machine learning},
  pages={325--333},
  year={2013},
  organization={PMLR}
}

@inproceedings{kim2019multiaccuracy,
  title={Multiaccuracy: Black-box post-processing for fairness in classification},
  author={Kim, Michael P and Ghorbani, Amirata and Zou, James},
  booktitle={Proceedings of the 2019 AAAI/ACM Conference on AI, Ethics, and Society},
  pages={247--254},
  year={2019}
}

@article{farh1991cultural,
  title={Cultural relativity in action: A comparison of self-ratings made by Chinese and US workers},
  author={FARH, JIING-LIH and Dobbins, Gregory H and CHENG, BOR-SHIUAN},
  journal={Personnel psychology},
  volume={44},
  number={1},
  pages={129--147},
  year={1991},
  publisher={Wiley Online Library}
}

@article{wuttke2024ai,
  title={AI conversational interviewing: Transforming surveys with LLMs as adaptive interviewers},
  author={Wuttke, Alexander and A{\ss}enmacher, Matthias and Klamm, Christopher and Lang, Max M and W{\"u}rschinger, Quirin and Kreuter, Frauke},
  journal={arXiv preprint arXiv:2410.01824},
  year={2024}
}

@article{chopra2023conducting,
  title={Conducting qualitative interviews with AI},
  author={Chopra, Felix and Haaland, Ingar},
  year={2023},
  publisher={CESifo Working Paper},
journal={CESifo Working Paper}
}

@article{leybzon2025ai,
  title={AI Telephone Surveying: Automating Quantitative Data Collection with an AI Interviewer},
  author={Leybzon, Danny D and Tirumala, Shreyas and Jain, Nishant and Gillen, Summer and Jackson, Michael and McPhee, Cameron and Schmidt, Jennifer},
  journal={arXiv preprint arXiv:2507.17718},
  year={2025}
}

@inproceedings{adhikari2025exploring,
  title={Exploring LLMs for Automated Generation and Adaptation of Questionnaires},
  author={Adhikari, Divya Mani and Hartland, Alexander and Weber, Ingmar and Cannanure, Vikram Kamath},
  booktitle={Proceedings of the 7th ACM Conference on Conversational User Interfaces},
  pages={1--23},
  year={2025}
}
\bibliographystyle{plainnat}

\appendix

\clearpage
\section{Appendix}

\section{Ethics Statement}
This work involves no human subjects or sensitive data (our training data is all from public sources), and we are unaware of any potential misuse, harm, or bias. No conflicts of interest or compromising sponsorships exist.

\section{Reproducibility}
We have made efforts to ensure the reproducibility of our results. We plan to release the full code publicly after publication. We \textit{currently} do not plan to release our finetuned LLM as a part of the release out of an abundance of caution regarding the uncertainty of licensing public data for LLM training, but will maintain it in an archive and allow other researchers access upon reasonable requests.

\subsection{Notation Table}
See Table~\ref{tab:variables} for definitions of all variables used in this paper

\section{Algorithm Design Notes}

For this section we use $s\in S$ to refer to text vectors from the set of all possible texts ($S$). These are represented as embeddings in practice. See Table~\ref{tab:variables} for the definition of each variable used in this paper.

\begin{table}[t]
    \centering
    \caption{Variables and definitions}\label{tab:variables}
    \begin{tabular}{rcl}
        \toprule
        \textbf{Variable} &\textbf{Type}& \textbf{Description}\\
        \midrule
        $j$ &$\mathbb{N}$& applicant ID/index\\
        $y$ &$\{0,1\}$& Ground truth, 1 if the applicant has the skill, 0 otherwise\\
        $\hat{y}$&$[0,1]$& Estimate of $y$ value gives probability of $y$ being 1\\
        $s()$  & $\str \rightarrow \str$ & Response simulator \\
        $q_\psi()$ & $\mathcal{X} \rightarrow \Delta(\mathcal{Q})$& question selector, produces a learned distributions over questions\\
        $\mathcal{Q}$ & & Set of questions, (pre-defined)\\
        $q_i$ & $\mathcal{Q}$& Specific question, indexed by number asked\\
        $s_i$ & $\str$& Specific response, indexed by number of responses\\
        $s_i^j$ & $\str$& Specific response, indexed by number of responses ($i$) and person responding ($j$)\\
        $i$ &$\mathbb{N}$&index for responses/questions to the same person\\
        $\mathcal{X}$ && Transcript (questions and responses) type\\
        $x_i$ &$\mathcal{X}$& Specific Transcript (questions and responses) up to response $i$\\
        $x_i^j$ &$\mathcal{X}$& Specific Transcript (questions and responses) up to response $i$ for person $j$\\
        $f_\phi()$ &$\mathcal{X} \rightarrow \mathbb{R}$ & score function, outputs score of candidate based on responses\\
        $f_\phi^*()$ & $\mathcal{X} \rightarrow \mathbb{R}$& score function, with multi-accuracy\\
        $\psi$ & & question selector weights (learned)\\
        $\phi$ & & score function weights (learned)\\
        $\operatorname{BERT}()$&$\str\rightarrow\mathbb{R}^n$ & Our pretrained embedding model, it maps text to vectors (\texttt{all-MiniLM-L6-v2})\\
        $\operatorname{ATTN}()$ &$\mathbb{R}^{n\times m}\rightarrow \mathbb{R}^{n}$&GPT style Attention mechanism, maps a sequences of vectors to single vector\\
        $Z$ &$\mathbb{R}$& Certainty, twice distance of $f(x_i^j)$ from $.5$\\
        $\mathcal{C}$ && Set of big 5 personality traits and constant function $\mathbf{1}$\\
        $c()$&$\mathcal{X} \rightarrow \mathbb{R}$&Big 5 personality trait score function (\texttt{MBTI}), pre-trained neural network, or $1$ for $c=\mathbf{1}$\\
        $\gamma$ & $\mathbb{R}$& Discount factor, set to 1 since we only care about the final evaluation\\
        $\alpha$ &$\mathbb{R}$&Question loss coefficient, controls how quickly we change $\psi$\\
        $\beta$ &$\mathbb{R}$&Score function loss coefficient, controls how quickly we change $\phi$\\
        $\epsilon$ &$\mathbb{R}$&Fairness constraint\\
        $l_c$ &$\mathbb{R}$&Calibration constants\\
        $\operatorname{softmax}()$&$\mathbb{R}^n\rightarrow\mathbb{R}^n$ & sigmoid function, maps a vector from $\mathbb{R}^n$ to a probability distribution, note $n$ can be 1\\
        $\sup()$&&Support\\
        $N$ &$\mathbb{N}$&Number of rollouts\\
        $M$ &$\mathbb{N}$&Rollout Depth\\
        $i_{\text{max}}$ &$\mathbb{N}$&Maximum number of questions\\
        $B$ &$\mathbb{N}$&Batch size\\
        $B^\text{fair}$ &$\mathbb{N}$&Batch size for fairness calibration\\
        $\delta$ &$\mathbb{R}$&The failure probability ($10^{-6}$)\\

        \bottomrule
    \end{tabular}
\end{table}

\subsection{Proof of Theorem~\ref{thm:ma}}\label{sec:supp-pf}
We first recall a lemma from \cite{gopalan2023loss}.
Let $\mathcal T$ denote the set of functions $g': \R \to \R$ such that: 1). $g'$ is continuous and monotonically increasing; 2). the range of $g'$ $ {\rm Im}(g')\supseteq [0,1].$
\begin{lemma} [Theorem~5.6 from~\cite{gopalan2023loss}]\label{thm:lemma0}
For a function $g$ and its derivative $g'\in\mathcal T$ whose range is [0,1], and let $l_g$ be its matching loss. Let $h^*$ be the optimal solution to the $\ell_1$-regularized loss minimization problem:
$$\min_{h \in Lin(C)}[\ell_g(\mathbf{y},h(\mathbf{x}))]+ \epsilon \sum_c |w_c|$$
where $h(x)=\sum_c w_c c(x)$.  Then the function $g' \circ h:\mathcal{X}\rightarrow [0,1]$ is a $(c,\epsilon)$-multiaccurate predictor.
\end{lemma}

Our setting is almost identical to that in Lemma~\ref{thm:lemma0}, except that our definition of $h(x)$ includes an offset term $\sigma^{-1}(f_\phi(x))$, which is independent of the optimization coefficients $l_c$.  The original proof of the lemma in~\cite{gopalan2021omnipredictors} goes through unchanged because the offset term disappears when computing $\frac{\partial h(x)}{\partial l_c}$.

By (the proof of) Lemma~\ref{thm:lemma0}, $f_\phi^*$ satisfies $(C,\epsilon)$-multiaccuracy:
\begin{align*}
\sup_{c\in\mathcal C} |\E_{j\in J}[c(x_i^j)\cdot (f_\phi^*(x_i^j)-y^j)]|\le \epsilon.
\end{align*}
Then, as for all $c\in\mathcal C$, we have $c\in[0,1]$, by Cauchy-Schwarz inequality, we have
\begin{align*}
\sup_{c\in\mathcal C} |\E_{j\in J}[c(x_i^j)\cdot (f_\phi(x_i^j)-y^j)]|\le& \sup_{c\in\mathcal C} |\E_{j\in J}[c(x_i^j)\cdot (f^*_\phi(x_i^j)-y^j)]|+\sup_{c\in\mathcal C} |\E_{j\in J}[c(x_i^j)\cdot (f_\phi(x_i^j)-f^*_\phi(x_i^j))]|\\
\le &\epsilon+\epsilon^*.
\end{align*}
This implies that $f_\phi^*$ is $(C,\epsilon+\epsilon^*)$-multiaccurate.

\section{Question Bank}\label{app:question_bank}

As discussed in Section~\ref{sec:results} we formulate our questions as hypotheticals where the subject must extrapolate to a world where they possess a specified skill. We do this due to the limitations of large language models, as we wish for all simulations to match our data distribution. The skills we consider as hypotheticals are based on the 30 most common in our dataset and are given in table~\ref{tab:questions}

\begin{table}[t]
\centering
\caption{Question bank skills}
\label{tab:questions}
\begin{tabular}{cl}
\toprule
\textbf{Rank} & \textbf{Skill} \\
\midrule
1  & Customer Service \\
2  & Microsoft Office \\
3  & Leadership \\
4  & Management \\
5  & Microsoft Excel \\
6  & Public Speaking \\
7  & Sales \\
8  & Microsoft Word \\
9  & Strategic Planning \\
10 & Project Management \\
11 & Social Media \\
12 & Marketing \\
13 & Team Building \\
14 & PowerPoint \\
15 & Research \\
16 & Training \\
17 & Event Planning \\
18 & Time Management \\
19 & Process Improvement \\
20 & Team Leadership \\
21 & Account Management \\
22 & Teamwork \\
23 & Program Management \\
24 & Sales Management \\
25 & Marketing Strategy \\
26 & Business Development \\
27 & Budgets \\
28 & New Business Development \\
29 & Social Media Marketing \\
30 & Strategy \\
\bottomrule
\end{tabular}
\end{table}

\section{Example Transcripts}

\begin{longtable}{p{3cm}p{12cm}}
\toprule
\textbf{Field} & \textbf{Content} \\
\midrule
\endfirsthead

\multicolumn{2}{c}
{{\bfseries \tablename\ \thetable{} -- continued from previous page}} \\
\toprule
\textbf{Field} & \textbf{Content} \\
\midrule
\endhead

\midrule
\multicolumn{2}{r}{{Continued on next page}} \\
\endfoot

\bottomrule
\endlastfoot

\multicolumn{2}{l}{\textbf{Example 1}} \\
\midrule
Prompt ID & 891004434 \\
Full Name & Laura Moore \\
Gender & female \\
Industry & Legal \\
Job Title & Administrative Coordinator \\
Company Size & 201-500 \\
Salary & \$35,000-\$50,000 \\
Skills & microsoft office, microsoft word, microsoft excel, customer service, time management, powerpoint, marketing \\
Prompt & I am a \\
Response & 27 year old professional woman who has successfully built a career in the business world. I enjoy the business world and being a part of that, and I have been doing that since I was 18 years old. I have a Bachelor of Arts in Political Science from San Francisco State University, and I have a Juris Doctorate from the University of California, Berkeley. \\
Model & llama \\
Max Tokens & 200 \\
\midrule

\multicolumn{2}{l}{\textbf{Example 2}} \\
\midrule
Prompt ID & 836422450 \\
Full Name & Emily Davis \\
Gender & female \\
Industry & Legal \\
Job Title & Administrative Coordinator \\
Company Size & 201-500 \\
Salary & \$35,000-\$50,000 \\
Skills & microsoft office, microsoft word, microsoft excel, customer service, time management, powerpoint, marketing \\
Prompt & I am a \\
Response & 2013 graduate of the University of California, Santa Barbara. I majored in Linguistics and minored in Spanish with the goal of one day becoming a lawyer or a teacher. I have since been accepted into the Masters of Legal Studies program at the University of Wisconsin-Madison. \\
Model & llama \\
Max Tokens & 200 \\
\midrule

\multicolumn{2}{l}{\textbf{Example 3}} \\
\midrule
Prompt ID & 1846149388 \\
Full Name & Amanda Williams \\
Gender & female \\
Industry & Marketing \\
Job Title & Marketing Assistant \\
Company Size & 51-200 \\
Salary & \$40,000-\$55,000 \\
Skills & social media, marketing, microsoft office, social media marketing, microsoft powerpoint, powerpoint, public speaking \\
Prompt & I am a \\
Response & 22-year-old who just graduated from the University of Arizona. I'm looking for a marketing position so I can gain experience and get my foot in the door. \\
Model & llama \\
Max Tokens & 200 \\
\midrule

\multicolumn{2}{l}{\textbf{Example 4}} \\
\midrule
Prompt ID & 1540984165 \\
Full Name & Michelle Wilson \\
Gender & female \\
Industry & Consulting \\
Job Title & Research Assistant \\
Company Size & 201-500 \\
Salary & \$45,000-\$60,000 \\
Skills & research, microsoft office, microsoft excel, teamwork, time management, powerpoint, public speaking \\
Prompt & I am a \\
Response & 2011 graduate from the University of Texas at Austin, majoring in Psychology and Neuroscience. I'm looking to further my career in the field of research, as well as pursue graduate studies in the area of psychology. \\
Model & llama \\
Max Tokens & 200 \\
\midrule

\multicolumn{2}{l}{\textbf{Example 5}} \\
\midrule
Prompt ID & 647743590 \\
Full Name & Emily Davis \\
Gender & female \\
Industry & Manufacturing \\
Job Title & Inventory Coordinator \\
Company Size & 501-1000 \\
Salary & \$40,000-\$55,000 \\
Skills & microsoft excel, microsoft office, process improvement, time management, teamwork, powerpoint, public speaking \\
Prompt & I am a \\
Response & 1.75 year college graduate with a degree in Applied Mathematics and Economics and an inventory coordinator position at an engineering and manufacturing company. \\
Model & llama \\
Max Tokens & 200 \\
\midrule
\end{longtable}

\section{Z-Scores}
\label{app:Z-scores}
\begin{figure}
    \centering
    \includegraphics[width=1\linewidth]{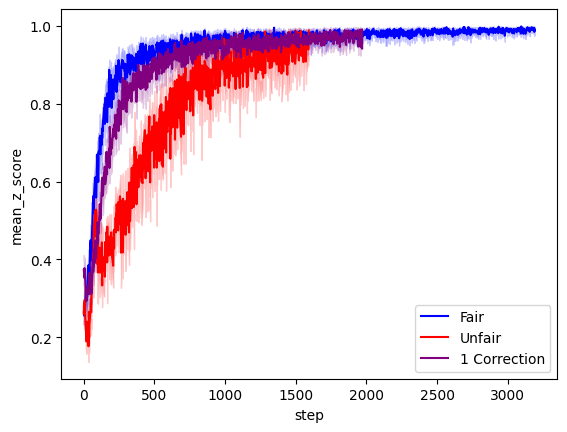}
    \caption{Z-scores on test samples during training with fairness correction at all epochs ({\color{blue}blue}), with the fairness correction applied only at step~$t$ ({\color{purple} purple}), and with the fairness correction never applied ({\color{red}red})}
    \label{fig:Z-scores}
\end{figure}
\end{document}